\documentclass[conference]{IEEEtran}
\IEEEoverridecommandlockouts

\usepackage{cite}
\usepackage{amsmath,amssymb,amsfonts}
\usepackage{algorithmic}
\usepackage{graphicx}
\usepackage{graphics}
\usepackage{textcomp}
\usepackage[export]{adjustbox}
\usepackage{multicol}
\usepackage{tabularx}
\usepackage{color}

\newcommand{\unaryminus}{\scalebox{0.75}[1.0]{\( - \)}}

\def\BibTeX{{\rm B\kern-.05em{\sc i\kern-.025em b}\kern-.08em
    T\kern-.1667em\lower.7ex\hbox{E}\kern-.125emX}}
\begin{document}

\title{
A Multi-component CNN-RNN Approach for Dimensional Emotion Recognition in-the-wild\\
}

\author{\IEEEauthorblockN{ Dimitrios Kollias}
\IEEEauthorblockA{\textit{Department of Computing} \\
\textit{Imperial College London}\\
United Kingdom \\
dimitrios.kollias15@imperial.ac.uk}
\and
\IEEEauthorblockN{Stefanos Zafeiriou}
\IEEEauthorblockA{\textit{Department of Computing} \\
\textit{Imperial College London}\\
United Kingdom \\}
\&\\
\IEEEauthorblockA{\textit{Center for Machine Vision and Signal Analysis} \\
\textit{University of Oulu}\\
Finland \\
s.zafeiriou@imperial.ac.uk}
}

\maketitle

\begin{abstract}
This paper presents our approach to the One-Minute Gradual-Emotion Recognition (OMG-Emotion) Challenge, focusing on dimensional emotion recognition through visual analysis of the provided emotion videos. The approach is based on a Convolutional and Recurrent (CNN-RNN) deep neural architecture we have developed for the relevant large AffWild Emotion Database. We extended and adapted this architecture, by letting a combination of multiple features generated in the CNN component be explored by RNN subnets. Our target has been to obtain best performance on the OMG-Emotion visual validation data set, while learning the respective visual training data set. Extended experimentation has led to best architectures for the estimation of the values of the  valence and arousal emotion dimensions over these data sets. This paper is a preliminary version of \cite{kollias14}.

\end{abstract}


\section{Introduction}
Automatic analysis of facial behaviour is the cornerstone of many application areas including Human Computer and Robot Interaction \cite{kollias10,kollia2009interweaving,simou2007fire}, Pervasive Computing, Ambient Intelligence, Virtual Reality etc. The research area of facial behaviour analysis includes problems such as the recognition of a set discrete universal facial expressions and their intensities, detection of the activated facial muscles, also known as Facial Action Units (FAUs) and estimation of their intensity, detection of micro-expressions (ME), estimation of facial affect in a continuous dimensional space (e.g., valence and arousal), as well as assist in flagging complex behavioral patterns such as deception, depression, autism, spectrum disorders \cite{tagaris1, tagaris2, kollias13}, schizophrenia etc. 

In order to facilitate research on the above problems the community has collected and annotated many databases (most of them in well-controlled conditions; nevertheless, recently the focus has turned into ”in-the-wild” data). In the beginning, data and annotations
were scarce, hence the community relied on extracting highly engineered handcrafted features and designing ad-hoc learning strategies. Naturally, as the data and annotations grew the community has recently started to capitalise on data-intensive technologies, such as deep learning. In this tutorial, motivated by the recent success of deep learning in computer vision and machine learning we survey the progress that has been recently made
on automatic understanding of facial behaviour using deep learning architectures. Based on the success of recent unified end-to-end trainable architectures we argue for a unified deep BehaviorNet \cite{kollias16} which can be trained in an end-to-end manner and simultaneously addresses (a) FAU activation detection, (b) recognition of discrete compound expressions and (c) estimation of valence and arousal. Finally we discuss about the challenges and possible future directions for facial behavior analysis based on deep-learning systems.

Estimation of continuous dimensions related to affect: according to the dimensional approach, affective behavior can be described by a number of latent continuous dimensions, that can better describe emotions experienced by humans in everyday life. The most commonly used dimensions include valence (indicating how positive or negative an emotional state is) and arousal (measuring the power of the emotion activation).

It is now widely accepted in both the computer vision and machine learning communities, that progress in a particular application domain is significantly catalysed when a large number of samples are collected in unconstrained conditions (also refferred to as "in-the-wild" data). Hence, for facial analysis we do not only need spontaneous behavior but the behavior should be captured in unconstrained conditions. To this end, recent efforts were made to collect and annotate facial behavior "in-the-wild". Example of databases annotated in terms of the: i) basic expressions, include the AFEW (audiovisual), AffectNet (static images), RAF-DB (static images), ii) action units, include the EmotioNet (static images), and (iii) valence-arousal, include the OMG-Emotion Dataset (audiovisual), Aff-Wild (audiovisual), SEWA (audiovisual), AffectNet (static images).   

Regarding the pipeline for facial behavior analysis, the standard paradigm was: (step 1) detect and/or track the face in an image sequence, (step 2) detect and/or track facial landmarks, (step 3) extract handcrafted features\footnote{Examples of handcrafted features include Histogram of Oriented Gradients (HoGs), Scale Invariant Feature Transform (SIFT), Local Binary Patterns (LBPs) and features from multiscale and multiorientation Gabor filterbanks  } , either around the landmarks or on the face region as a whole and (step 4) use the features, the landmarks and data augmentation techniques \cite{kollias8,kollias9} in classification/regression using affective labels. Recently this paradigm has shifted from utilizing handcrafted features to utilizing features learned from DCNNs and/or RNNs, a shift motivated by the striking performance achieved when utilizing DNNs for several computer vision tasks.

In this paper, we address the issue of estimating  valence and arousal utilizing the One-Minute-Gradual  Emotion  Dataset  (OMG-Emotion Dataset) and focusing only on the visual modality. Note that this paper is a preliminary version of \cite{kollias14}. We present our approaches and submissions to the OMG-Emotion Challenge.
The rest of this paper is organized as follows.
Section \label{s2} reviews the related work and existing  state-of-the-art methods for facial expression recognition with emphasis on the dimensional model of affect. Section \label{s3} gives a brief description of databases used in our experiments, that were the OMG-Emotion Dataset and the Aff-Wild database. Section \label{s4} presents our methodologies; the created novel deep neural networks, including ensembles and fusion of networks, targeted for valence-arousal estimation. Section \label{s5} provides an evaluation of our approach by presenting the performance results and doing comparisons. Finally, Section \label{s6} concludes the paper.

Generating emotion databases, especially in-the-wild, and organizing related Challenges has recently proven to be of great significance in the development of novel emotion recognition techniques \cite{kollias11} and methodologies \cite{kollias12}. Such Challenges included, in 2017, the EmotiW \cite{dhall2017individual}, focusing on categorical emotion recognition in-the-wild, the AVEC \cite{ringeval2017avec}, focusing on dimensional emotion recognition and the AffWild \cite{kollias1,kollias3}  Challenge, which the authors of this paper developed, focusing on dimensional emotion recognition in-the-wild. 

The current One-Minute-Gradual Emotion Dataset (OMG-Emotion Dataset) \cite{barros2018omg} is composed of 420 relatively long emotion videos with an average length of 1 minute, collected from a variety of Youtube channels. It provides both dimensional and categorical annotations, as well as textual descriptions of what was spoken in the videos. 

In this paper we focus only on the visual part of the provided videos. Contextual information can be combined with our approach to improve the obtained results, but this is rather an extension, of our target in the presented approach. Our approach is based on a CNN-RNN deep neural architecture which has produced best performance on the AffWild database \cite{kollias2,kollias3}. We use the specific architecture to initialize the training procedure \cite{kollias3}, using the OMG-Emotion visual training dataset. We then extend and adapt this architecture, by combining multiple components in the deep network structure, the error criteria used in network training and applying different post-processing strategies. We evaluate the performance of the developed approaches on the OMG-Emotion validation dataset and present the obtained results. These approaches have been also applied to the OMG-Emotion test dataset.    

Section II gives a short description of both AffWild and OMG-Emotion Databases. The proposed approach is presented in Section III. The experimental study is summarized in Section IV. Section V presents the derived conclusions and possible future extensions.

\section{Dimensional Emotion Databases in-the-wild}

In this Section we provide a short description of the OMG-Emotion database, analysis of which is the target of this paper, as well as of the AffWild Database, which has been used to initialize the deep neural network architecture developed in this paper.  

\subsubsection{AffWild Database}

We have created the Aff-Wild database \cite{kollias1,kollias2,kollias3} consisting of 298 videos, with a total
length of more than 30 hours. The aim was to collect spontaneous
facial behaviors in arbitrary recording conditions. To
this end, the videos were collected using the Youtube video
sharing web-site. 
The main keyword that was used to retrieve
the videos was ”reaction”. The database displays subjects reacting to a variety of stimuli (i.e., from a video to tasting
something hot or disgusting). The subjects display
both positive or negative emotions (or combinations of
them). The videos contain subjects from different genders and ethnicities with high variations in head pose and lightning. Different annotators have provided valence and arousal emotion dimension values for all videos in the database. Let us note that Aff-Wild has been extended to Aff-Wild2 \cite{kollias15,kollias4,kollias5} that contains more videos and annotations for valence-arousal, action units and basic expressions.

\subsubsection{OMG-Emotion Dataset}

The dataset contains in-the-wild videos where emotion expressions emerge and develop over time based on monologued scenarios, annotated by a large number of annotators. In particular, the dataset includes a total of 178 unique videos, totalizing 2725
video clips with each clip consisting of a single utterance.
Each video is split into several sequential utterances, each one
with an average length of 8 seconds, and having an average
length of around 1 minute. After annotation collection, a total of 11635 unique annotations were obtained, averaging 5
annotations per video. Youtube was used as the source for dataset samples, with samples containing ample variety
in recording conditions, speakers and expressed emotions.
Acted monologues can have utterances displaying
both neutral and salient emotions, depending on the setting
of the monologue itself.

\section{The Proposed Approach}\label{nets}
  
For this challenge we experimented with many different architectures, both CNN and CNN plus RNN. The design of our models was based on two widely used CNN architectures: VGG-16 and ResNet-50.

\subsection{Exploiting VGG Face/VGG-16 CNN networks}\label{nets_vgg_cnn}

\mbox{}\\
Table \ref{VGG-16} shows the configuration of the CNN architecture based on VGG-16. It is composed of 8 blocks. For each convolutional layer the parameters are denoted as (channels, kernel, stride) and for the max pooling layer as (kernel, stride).  The output number of units is also shown in the Table. The use of 2 Fully-Connected (FC) layers, before the final output layer, was found to provide the  best results. 

\begin{table}[h]
\caption{CNN architecture based on VGG-16}
\label{VGG-16}
\centering
\begin{tabular}{|c|c|c|}
\hline
block 1 & $2  \times$ conv layer & (64, $3 \times 3$, $1 \times 1$) \\
&$1 \times$ max pooling & ($2 \times 2$, $2 \times 2$) \\
\hline
block 2 & $2  \times$ conv layer & (128, $3 \times 3$, $1 \times 1$) \\
&$1 \times$ max pooling & ($2 \times 2$, $2 \times 2$) \\
\hline
block 3 & $3 \times$ conv layer & (256, $3 \times 3$, $1 \times 1$) \\
&$1 \times$ max pooling & ($2 \times 2$, $2 \times 2$) \\
\hline
block 4 & $3 \times$ conv layer & (512, $3 \times 3$, $1 \times 1$) \\
&$1 \times$ max pooling & ($2 \times 2$, $2 \times 2$) \\
\hline
block 5 & $3 \times$ conv layer & (512, $3 \times 3$, $1 \times 1$) \\
&$1 \times$ max pooling & ($2 \times 2$, $2 \times 2$) \\
\hline
block 6 &fully connected 1 & 4096\\
\hline
block 7 &fully connected 2 & 2048\\
\hline
block 8 &fully connected 3 & 2\\
\hline
\end{tabular}
\end{table}

\subsection{Developing CNN plus RNN architectures}

\subsubsection{CNN plus RNN, based on VGG-16}\label{nets_vgg_cnn_rnn}

\mbox{}\\

In order to match the continuous evolution of the contextual information in the data, we developed and used CNN-RNN architectures. Table \ref{rnn} shows the configuration of the developed basic CNN-RNN. The CNN part of this architecture is based on the convolutional and pooling layers of VGG-16. It is followed by a fully connected layer. On top of that, a 2 layered RNN is stacked, giving the final estimates for valence and arousal.

\begin{table}[h]
\caption{CNN-RNN architecture based on convolution and pooling layers of previously described CNNs. }
\label{rnn}
\centering
\begin{tabular}{|c|c|c|}
\hline
block 1 & VGG-16 conv \& pooling parts & \\
\hline
block 2 &fully connected 1 & 4096  \\
&dropout layer&\\
\hline
block 3 & RNN layer 1 & 128\\
&dropout layer&\\
\hline
block 4 & RNN layer 2 & 128\\
\hline
block 5 &fully connected 2 & 2\\
\hline
\end{tabular}
\end{table}

In our approach we developed three extensions of this architecture, by extracting more features from the CNN part and exploiting them with different RNN subnets.  
In the first of these extensions, the CNN-1RNN configuration was also based on the convolutional and pooling layers of VGG-16 followed by a fully connected one with 4096 units. In this, however, the outputs of: i) the last convolutional layer of VGG-16, ii) the last pooling layer of VGG-16 and iii) the fully connected layer, were concatenated and given as input to a 2 layered RNN stacked on top. Then the output layer followed, that gave the final estimates.

The second extended CNN-2RNN configuration was based on the convolutional and pooling layers of VGG-16 followed by a fully connected one with 4096 units, as well. 2 RNNs were stacked on top of it: i) the output of the last pooling layer of VGG-16 was given as input to one of them and ii) the output of the fully connected layer was given as input to the other RNN. Then, the outputs of the RNNs were concatenated and were passed to the output layer that gave the final estimates.  
A similar configuration was tested, including a hidden layer before the output, so as to exploit more complex correlations between the outputs of the above two RNNs (CNN-2RNN-FC).

The last extended CNN-3RNN configuration was also based on the convolutional and pooling layers of VGG-16 followed by a fully connected one with 4096 units. However, 3 RNNs were stacked on top of it:  i) The output of, either the last or the one before the last (penultimate), convolutional layer of VGG-16 was given as input to one RNN, ii) the output of the last pooling layer of VGG-16 was given as input to the second RNN and iii) the output of the fully connected layer was given as input to the third RNN. Then, the outputs of the RNNs were concatenated and were passed to the output layer that gave the final estimates.  
A similar configuration, with a hidden layer before the output, was tested as well, so as to exploit more complex correlations between the outputs of the afore mentioned 3 RNNs (CNN-3RNN-FC).\\

\begin{figure*}[t]
\centering
\adjincludegraphics[height=9cm,trim={.0\width} {.0\totalheight} {.0\width} {.0\totalheight}]{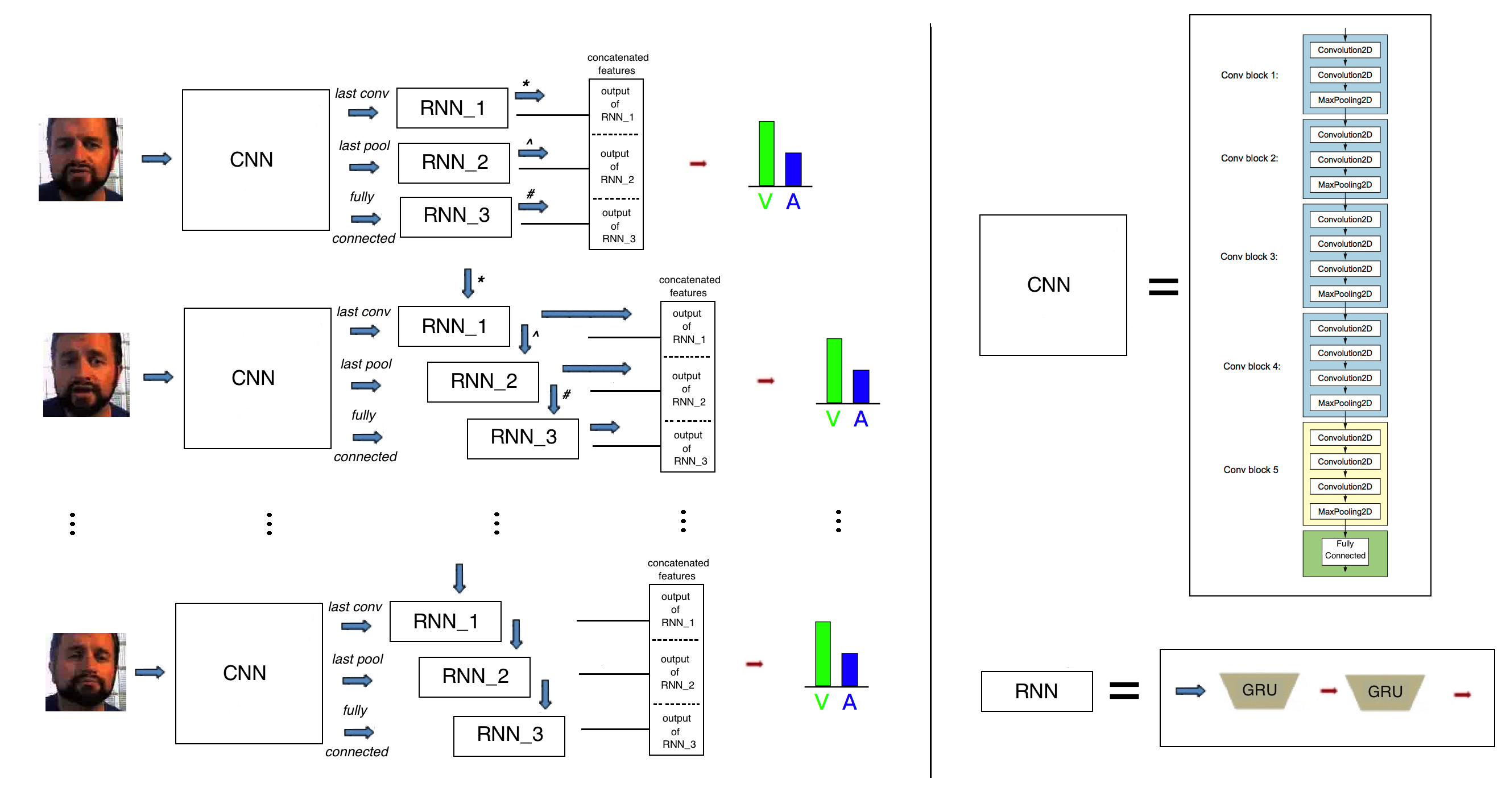} 
\caption{The network architecture that gave the best results for valence}
\label{cnn_3rnn}
\end{figure*}

\begin{figure}[t]
\centering
\adjincludegraphics[height=7cm,trim={.03\width} {.0\totalheight} {.03\width} {.0\totalheight}]{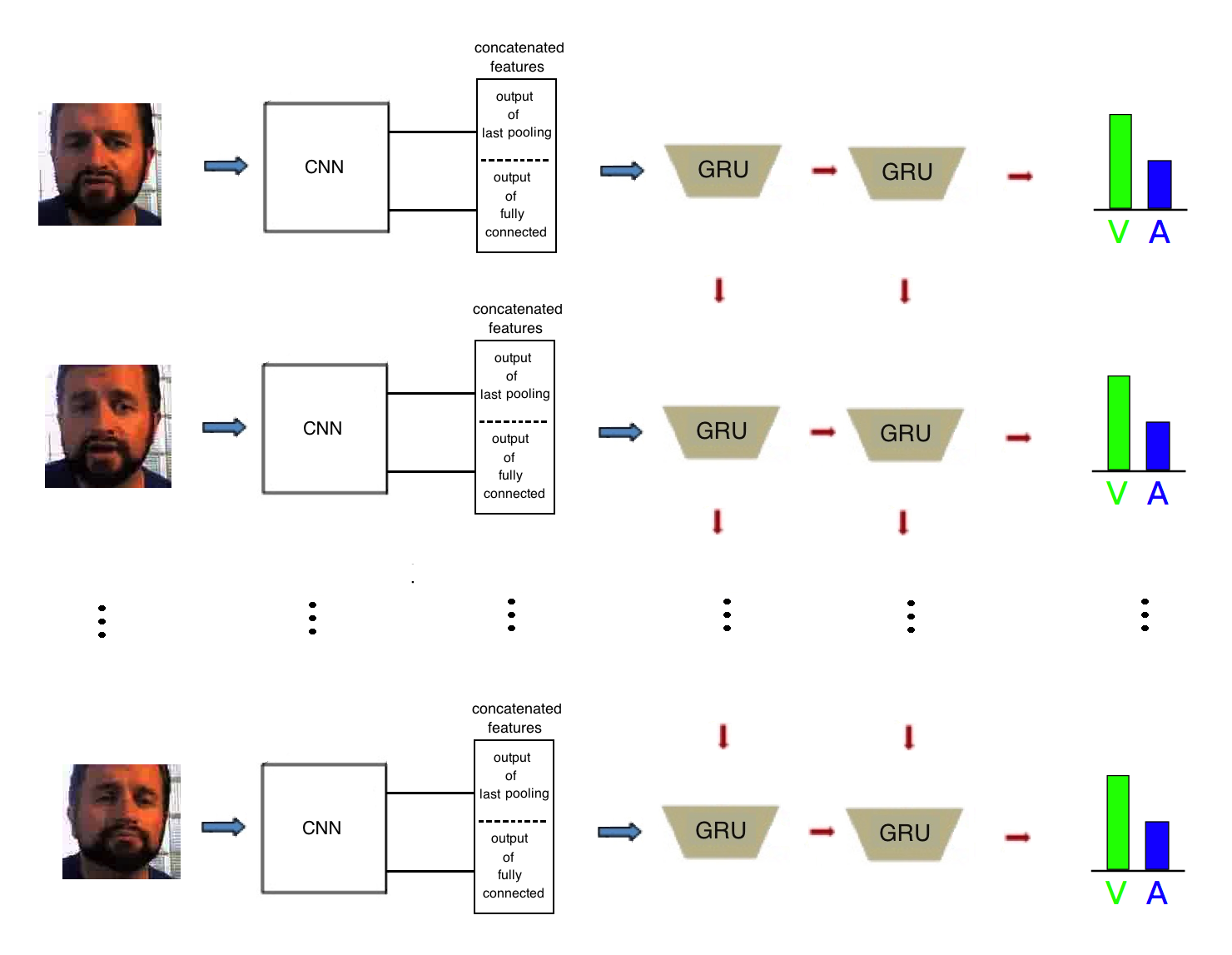} 
\caption{The network architecture that gave the best results for arousal}
\label{cnn_1rnn}
\end{figure}

\subsubsection{CNN plus RNN, based on ResNet-50}\label{nets_resnet_cnn}

\mbox{}\\

We also tested CNN-RNN architectures, based on the deep CNN residual network (ResNet) consisting of 50 layers \cite{he2016deep}.
The first layer of the ResNet-50 model is comprised of a $7 \times 7$ convolutional layer with 64 feature maps, followed by a max pooling layer of size $3 \times 3$. Next, there are 4-bottleneck blocks, where after each block a shortcut connection is added. Each of these blocks is comprised of  3 convolutional layers of sizes $1 \times 1$, $3 \times 3$, and $1 \times 1$ with different number of feature maps. 

As in the VGG case, in addition to the use of ResNet-50, we stacked, on top, a 2-layered RNN that provided the final estimates. We also experimented with the use of a fully connected layer before the RNN (CNN-FC-RNN) .\\

\subsubsection{Fusion of Networks / Ensemble methodology}\label{nets_fusion}

\mbox{}\\

The last methodology that we followed was based on the fusion of some of the above described networks. The fused network consisted of two different CNN-RNN architectures, of which, the outputs of the RNN layers were concatenated and passed either to : i) the output layer, or 
ii) a fully connected layer that preceded the output layer.
In more detail, we fused: i) a CNN-RNN network, where the CNN part was based on the convolutional and pooling layers of VGG-16 followed by a fully connected one with 4096 units and the RNN consisted of 2 layers and ii) another CNN-RNN network, where the CNN part was based on the convolutional and pooling layers of ResNet-50 followed by either: i) a RNN, or ii) a fully connected one and a RNN. The RNN in both cases had 2 layers.

\subsection{Pre-processing, Network Training and Testing}

We cropped the input images to the bounding boxes \cite{avrithis2000broadcast}. In order to do so, we used the Deformable Part Model (DPM) detector ffld2 \cite{felzenszwalb2010object} \cite{mathias2014face}. This resulted in generation of the facial images to be processed. Then, we resized those facial images to $96 \times 96 \times 3$  pixel resolution and normalized the images' intensity values to the range $[-1, 1]$. Those images were then fed as input to our networks for training. The training was end-to-end. Before training on the OMG-Emotion training Dataset, we pretrained our networks on the large AffWild database. In the case of model fusion, we tried two different techniques: either we pretrained the models on the AffWild and then trained them together on the OMG-Emotion training Dataset or we used the weights of the networks that had the best performance when trained alone on the OMG-Emotion training Dataset and then trained them together.

To train our CNN plus RNN networks, we used a batch size of 4 and sequence length of 80 consecutive frames. The annotation was made per utterance, meaning that one value for valence and one for arousal was given to each utterance. For training our networks, each sequence was given the value of the utterance that it came from.
For testing the performances, the CCC measure was used as evaluating criterion and each utterance was assigned the mean or median of the values of the frames that it consisted of.

\subsection{Objective function}

Our loss function was based on the Concordance Correlation Coefficient (CCC) metric that has been shown to provide better insight on whether the prediction follows the structure of the ground truth annotation. Hence, our loss was defined as $\mathcal{L}_c = 1 - \rho_c$, with:

\begin{equation} \label{eq:1}
\rho_c = \frac{2 s_{yf(x)}}{s_{f(x)}^2 + s_y^2 + (\bar{f(x)} - \bar{y})^2}
\end{equation}

\noindent
where $s_y$ and $\bar{y}$ are the variance and the corresponding mean value of the ground truth labels, $s_{f(x)}$ and $\bar{f(x)}$ are the variance and the corresponding mean (across batches) of ($f(x)$); with x being the predictions of the network across a sequence and the function $f(.)$ computing either the mean or the median 
; $s_{yf(x)}$ is the respective covariance value.

In other words, we first computed the mean or median of the predictions of our network for each sequence and then we used those values (in every iteration those values were 4 since the batch size was 4) to compute the CCC.

\subsection{Post-processing}

Finally, for all investigated methods, a chain of post-processing steps was applied: i) to the per frame predictions, median filtering (with size of window being 81 for valence and 3 for arousal) and ii) to the per utterance predictions, smoothing (especially to those that consisted of too few frames). Any of these post-processing steps was kept when an improvement was observed on the CCC of the validation set, and applied then, with the same configuration on the test partition.

\section{Experimental Results and Ablation Study}

In the following we provide specific information on the selected structure and parameters of the developed end-to-end neural architectures and also experimental results for most of them. 

\subsubsection{VGG-16 CNN network}

Extensive experiments have been performed by selecting different network parameter values, including: (1) the number of fully connected layers (1 $\unaryminus$ 3; optimal was found to be 2); (2) 
the number of neurons in each fully connected layer (1st FC: 1024 $\unaryminus$ 4096; optimal was found to be 4096 ; 2nd FC: 1024 $\unaryminus$ 4096; optimal was found to be 2048); (3) the batch size used for network parameter updating (10 $\unaryminus$ 100; optimal was found to be 80); (4) the value of the learning rate (0.0001 $\unaryminus$ 0.001; optimal was found to be 0.001); (5) the dropout probability value (0.40 $\unaryminus$ 0.8; optimal was found to be 0.5).

\subsubsection{CNN plus RNN, based on VGG-16}

For the part before the RNN layers, we kept the convolutional and pooling parts of VGG-16 followed by a fully connected layer with 4096 hidden units and dropout probability equal to 50. Above that, we stacked either one or two or three RNNs as described in Section \ref{nets_vgg_cnn_rnn}. In some architectures those RNNs were followed by a fully connected layer (before the output); after experimenting with the number of units in the range 32 $\unaryminus$ 256, we found that the best results were obtained with 64 units. For the RNNs, the Gated Recurrent Unit (GRU) neuron model was preferred to the Long Short Term Memory (LSTM) one and a 2-layered architecture as well. After extensive experimentation, dropout probability equal to 0.2, between the 2 GRU layers, was found to give optimal results.

\subsubsection{CNN plus RNN, based on ResNet-50}

Here we experimented with putting a fully connected layer after the ResNet-50 and before the RNN. We selected hidden units in the range 0 $\unaryminus$ 4096. Optimal results have been found when we did not have a fully connected layer.

\subsubsection{Fusion of Networks / Ensemble methodology}

Here we experimented with putting (or not) a fully connected layer before the output. We selected hidden units in the range 0 $\unaryminus$ 128. Optimal results have been found when we did not have a fully connected layer.

\begin{table}[!h]
\caption{best CCC evaluation of valence \& arousal predictions on the validation OMG-Emotion Dataset provided by the networks based on VGG-16, described in Sections \ref{nets_vgg_cnn} \& \ref{nets_vgg_cnn_rnn}}
\label{vgg16_rnn}
\centering
\begin{tabular}{ |c||c|c| }
\hline
\multicolumn{1}{|c||}{} & \multicolumn{2}{|c|}{CCC} \\
\hline
     & Valence & Arousal  \\
\hline
CNN only & 38.1 & 25.5  \\
 \hline
basic CNN-RNN & 44.7  & 26.9  \\
 \hline
CNN-1RNN & 40.74 & \textbf{30.05}  \\
 \hline
CNN-2RNN & 42.74 & 28.61  \\
 \hline
CNN-2RNN-FC & 42.11 & 28.33  \\
 \hline
CNN-3RNN-2nd-last-conv & 45.34 & 21.5  \\
 \hline
CNN-3RNN-last-conv & \textbf{45.6} &  24.6 \\
 \hline
\end{tabular}
\end{table}

\begin{table}[!h]
\caption{best CCC evaluation of valence \& arousal predictions on the validation OMG-Emotion Dataset provided by the networks based on ResNet-50, described in Sections \ref{nets_resnet_cnn}}
\label{resnet_rnn}
\centering
\begin{tabular}{ |c||c|c| }
\hline
\multicolumn{1}{|c||}{} & \multicolumn{2}{|c|}{CCC} \\
\hline
     & Valence & Arousal  \\
\hline
CNN only & 38.4 & 21.64  \\
 \hline
basic CNN-RNN & \textbf{44.4} & \textbf{26.7}  \\
 \hline
CNN-FC-RNN &  40.7  & 26.4  \\
 \hline
\end{tabular}
\end{table}

\begin{table}[!h]
\caption{best CCC evaluation of valence \& arousal predictions on the validation OMG-Emotion Dataset provided by the fused networks, as described in Sections \ref{nets_fusion}}
\label{fused}
\centering
\begin{tabular}{ |c||c|c| }
\hline
\multicolumn{1}{|c||}{} & \multicolumn{2}{|c|}{CCC} \\
\hline
     & Valence & Arousal  \\
\hline
\begin{tabular}{@{}c@{}} VGG-16-FC-RNN \\ + \\ ResNet-50-RNN \\ + \\ Output layer \end{tabular}  & \textbf{45.47}  & \textbf{27.47}  \\
 \hline
\begin{tabular}{@{}c@{}} VGG-16-FC-RNN \\ + \\ ResNet-50-FC-RNN \\ + \\ Output layer \end{tabular}  & 44.1  &  27.3 \\
 \hline
\begin{tabular}{@{}c@{}} VGG-16-FC-RNN \\ + \\ ResNet-50-RNN \\ + \\ FC \\ + \\ Output layer \end{tabular}  & 45.3  & 25.07  \\
 \hline
\begin{tabular}{@{}c@{}} VGG-16-FC-RNN \\ + \\ ResNet-50-FC-RNN \\ + \\ FC \\ + \\ Output layer \end{tabular}  & 45  & 24.5  \\
 \hline

\end{tabular}
\end{table}

\begin{table}[!h]
\caption{CCC evaluation of valence \& arousal predictions on the validation OMG-Emotion Dataset, after applying post-processing techniques, provided by the best networks described in Tables \ref{vgg16_rnn}, \ref{resnet_rnn} \& \ref{fused}}
\label{post_process}
\centering
\begin{tabular}{ |c||c|c| }
\hline
\multicolumn{1}{|c||}{} & \multicolumn{2}{|c|}{CCC} \\
\hline
     & Valence & Arousal  \\
\hline
VGG-16-1RNN & 43.1 & \textbf{31.1}  \\
 \hline
VGG-16-3RNN-last-conv & \textbf{49.1} & 26.1  \\
 \hline
classic ResNet-RNN & 46.2 & 28.1  \\
 \hline
\begin{tabular}{@{}c@{}} VGG-16-FC-RNN \\ + \\ ResNet-50-RNN \\ + \\ Output layer \end{tabular} & 48.45   & 28.86  \\
 \hline
\end{tabular}
\end{table}

Tables \ref{vgg16_rnn}-\ref{post_process} present the best results obtained by the developed architectures. In particular, Table \ref{vgg16_rnn} shows that the best results obtained by VGG-16 based networks were provided by CNN-1RNN (for arousal) and CNN-3RNN (for valence).  Table \ref{resnet_rnn} shows the best results obtained by ResNet-50 based networks which are a bit lower than those of VGG-16. Table \ref{fused} shows the best results obtained by the fusion approach. Table \ref{post_process} shows the best results obtained after applying post-processing to the best networks in the above Tables. It can be seen that with post-processing the best results have been improved by 7.7$\%$ for valence and 3.5$\%$ for arousal.

\section{Conclusions}

In this paper we have proposed a multi-component CNN-RNN architecture for dimensional - in terms of valence and arousal - emotion recognition based on facial expression analysis in the provided OMG-Emotion visual datasets. The obtained best CCC performances in the validation datasets were quite high, 0.49 for valence and 0.31 for arousal, much higher than the provided baselines of 0.23 and 0.12 respectively. Respective submissions have been made for the OMG-Emotion test dataset.

\bibliographystyle{spmpsci}      
\bibliography{sample}   

\end{document}